\documentclass[conference]{IEEEtran}
\IEEEoverridecommandlockouts
\def\BibTeX{{\rm B\kern-.05em{\sc i\kern-.025em b}\kern-.08em
    T\kern-.1667em\lower.7ex\hbox{E}\kern-.125emX}}
\usepackage{cite}
\usepackage{amsmath,amssymb,amsfonts}
\usepackage{algorithmic}
\usepackage[numbers,square,sort&compress]{natbib}
\usepackage{textcomp}
\usepackage{graphicx}
\usepackage{xcolor}

\usepackage{hyperref}
\usepackage{enumitem}
\usepackage{colortbl}
\usepackage{wrapfig}
\usepackage{pifont}    %
\usepackage{booktabs}
\usepackage{siunitx}
\usepackage{multirow}
\usepackage{makecell}
\usepackage{caption}
\usepackage{subcaption}
\newcommand{\xmark}{\ding{55}}
\usepackage[most]{tcolorbox}
\definecolor{black}{HTML}{000000}
\definecolor{CreamyOffWhite}{HTML}{F9E4D9}
\definecolor{SoftRoseTaupe}{HTML}{C99F99}
\definecolor{WarmMochaMousse}{HTML}{A47864}
\definecolor{DeepChocolate}{HTML}{8D6344}
\definecolor{LightTeal}{HTML}{E0F7F9}   %
\definecolor{DeepNavy}{HTML}{003366}    %
\definecolor{Black}{HTML}{000000}
\definecolor{CustomLightRed}{HTML}{F3D7D5}
\definecolor{CustomRed}{HTML}{C30E00}
\definecolor{CustomLightGreen}{HTML}{DFF2D9}
\definecolor{CustomGreen}{HTML}{40B814}

\newcommand{\rairebuttal}[1]{\textcolor{black}{#1}}

\def\rebuttal#1{\textcolor{black}{#1}}

\definecolor{softyel}{RGB}{255,241,184} %

\newcommand{\hlc}[1]{\cellcolor{CustomLightGreen}{#1}}

\begin{document}

\title{Scaling Short-Term Memory of Visuomotor Policies for Long-Horizon Tasks
    \author{
    \IEEEauthorblockN{
        Rutav Shah$^{1,2}$,
        Rajat Kumar Jenamani$^{1}$,
        Xiaohan Zhang$^{1}$,
        Lingfeng Sun$^{1}$,\\
        Roberto Martín-Martín$^{2}$,
        Yuke Zhu$^{2}$,
        Deva Ramanan$^{1}$,
        Karl Schmeckpeper$^{1}$
    }
    \IEEEauthorblockA{$^1$Robotics and AI Institute, $^2$The University of Texas at Austin}
    }
}

\maketitle

\begin{abstract}
Many robotic tasks require short-term memory, whether it’s retrieving an object that's no longer visible or turning off an appliance after a set period.
Yet, most visuomotor policies trained via imitation learning remain myopic, relying only on immediate sensory input without using past experiences to guide decisions.
We present PRISM, a transformer-based architecture for visuomotor policies to effectively use short-term memory via two key components: (i) gated attention, which selectively filters retrieved information to suppress irrelevant details, improving performance by reducing the spurious correlations between the history and current action prediction, and (ii) a hierarchical architecture that first compresses local interactions into compact tokens and then integrates them to capture temporally extended dependencies, improving its compute and memory footprint.
Together, these mechanisms enable us to scale short-term memory in visuomotor policies for up to two minutes.
To systematically evaluate memory in visuomotor control, we introduce \textsc{ReMemBench}---a benchmark of eight diverse household manipulation tasks spanning four categories of short-term memory---designed to foster general memory mechanisms rather than siloed, task-specific solutions.
\textsc{PRISM} consistently outperforms prior works, including transformer-based visuomotor policies with short-term memory, recurrent architectures, and other transformer variants---achieving an absolute improvement of $5\%$--$12\%$ over the strongest baseline in \textsc{ReMemBench} and real-world evaluations.
On the standard \textsc{RoboCasa} and LIBERO benchmarks, it achieves absolute improvements of $11\%$--$15\%$ over its no-memory variant and fine-tuned Vision-Language-Action baselines such as GR00T-N1-3B and OpenVLA, despite not leveraging any large-scale pretraining.
Together, PRISM and \textsc{ReMemBench} establish a foundation for developing and evaluating short-term memory--augmented visuomotor policies that scale to long-horizon tasks.
Additional materials are available at~\href{https://shahrutav.github.io/short-term-memory}{\url{shahrutav.github.io/short-term-memory}}
\end{abstract}
\section{Introduction}
Memory is central to intelligent behavior: human cognition depends on retaining past experiences to guide ongoing actions~\citep{squire2004memory,anderson2015cognitive}.
Classic studies in human cognition distinguish between sensory, short-term, and long-term memory~\citep{atkinson1968human}.
Among these, short-term memory, which spans seconds to minutes, is essential in everyday activities such as remembering to stir at regular intervals, retrieving objects that are no longer visible, or counting scoops of salt~\citep{baddeley2020working}.
Without it, behavior collapses to responses grounded only in the present observation, unable to account for past experiences.

\begin{figure*}[t]
    \centering
    \includegraphics[width=0.95\textwidth,trim=0pt 0pt 0pt 0pt,clip]{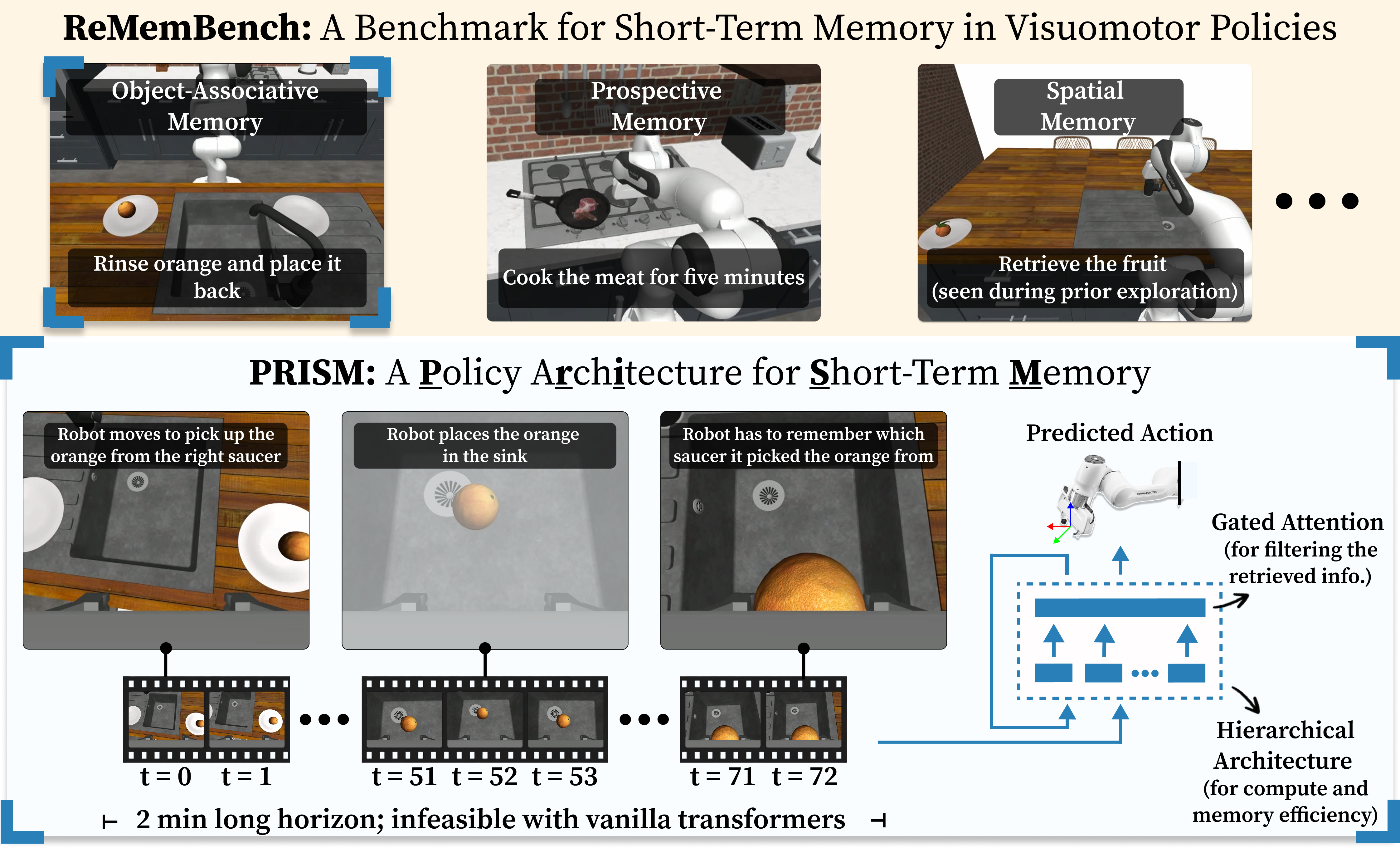}
    \vspace{-0.3cm}
    \caption{We propose \textsc{ReMemBench}, a benchmark for evaluating visuomotor policies with short-term memory (top), and PRISM, an architecture for visuomotor policies with gated attention and a hierarchical architecture, enabling robots to use short-term memory for partially observable long-horizon tasks (bottom).}
    \label{fig:teaser}
\end{figure*}

Recent advances in robot learning have led to the development of generalist visuomotor policies through imitation learning~\citep{black2024pi0,bjorck2025gr00t,shukor2025smolvla}.
However, they do not target imitation learning under partial observability and consequently adopt policies without short-term memory, relying solely on immediate sensory inputs.
Most visuomotor policies rely on vanilla transformer backbones~\citep{vaswani2017attention}, and naively extending this architecture with short-term memory via longer context windows poses two main drawbacks.
\rairebuttal{
First, it is susceptible to irrelevant information in the context window~\citep{hong2025context}.
This can induce spurious correlations, \textit{e.g.}, linking past distractor locations to a left or right placement decision, degrading test performance~\citep{de2019causal,wen2020fighting}.
While these spurious correlations could be reduced with more data, scaling robot data is challenging because teleoperating robots is labor-intensive.
Second, at test time, its computational and memory footprint scale linearly with the number of tokens in the context length, making it inefficient to process high-dimensional sensory inputs, such as image observations, over long temporal sequences~\citep{vaswani2017attention}.}
These drawbacks limit the scalability of short-term memory over long horizons in visuomotor policies.

To bridge this gap, we introduce \textbf{PRISM}: a \textbf{\underline{P}}olicy A\textbf{\underline{r}}ch\textbf{\underline{i}}tecture with \textbf{\underline{S}}hort-Term \textbf{\underline{M}}emory for visuomotor control that maps sequences of past sensory inputs and actions to future actions in a closed-loop setting.
Building on technical innovations in related fields of vision and language processing, PRISM extends the vanilla transformer architecture with two key modifications.
First, a post-attention gating mechanism selectively filters information, thereby introducing an explicit pathway to suppress irrelevant information~\citep{qiu2025gated}.
Second, a hierarchical design compresses local information into compact tokens and models long-horizon dependencies by attending over this compact set.
This design factors a single expensive computation into two cheaper stages, substantially reducing computational cost~\cite{bertasius2021space,arnab2021vivitvideovisiontransformer}.
Moreover, because future-timestep computation depends only on a compact set of tokens, PRISM stores this reduced set at test time, thereby reducing the memory footprint.
By combining information filtering with a resource-efficient hierarchical architecture, PRISM scales short-term memory to sequences of up to two minutes at $2$--$3$ Hz, an order of magnitude larger than prior approaches~\citep{torne2025learning}.

To systematically evaluate visuomotor policies with short-term memory, we introduce~\textbf{\textsc{ReMemBench}}, a benchmark built upon \textsc{RoboCasa}~\citep{nasiriany2024robocasa}, inspired by short-term memory categories in cognitive science (\autoref{fig:teaser}).
Existing embodied AI benchmarks~\citep{pasukonis2022evaluating, fang2025sam2act} either probe memory indirectly or focus narrowly on a single or selective aspects, such as spatial memory~\citep{fang2025sam2act}, making it difficult to assess memory capabilities holistically.
~\textsc{ReMemBench} instead spans four categories of short-term memory---spatial, prospective, object-associative, and object-set---across eight visuomotor tasks (\autoref{sec:main:benchmarking}).
By unifying these categories into a single benchmark, \textsc{ReMemBench} encourages the development of general memory mechanisms rather than siloed task-specific solutions.

On \textsc{ReMemBench}, PRISM outperforms a wide range of strong baselines---including recurrent, transformer-based, and specialized policy architectures for short-term memory---by an absolute margin of $5\%$.
Moreover, by augmenting the state with recent observations, PRISM can resolve action ambiguity under visually identical inputs (\textit{e.g.}, ``just grasped'' vs. ``about to place''), reducing multimodality and improving performance on benchmarks not designed to test memory
In particular, it shows an absolute improvement of $11\%$--$12\%$ on \textsc{RoboCasa} and LIBERO relative to their no-memory variants.
Notably, it outperforms strong pretrained models with an absolute improvement of $11\%$ over GR00T-N1-3B~\citep{bjorck2025gr00t} finetuned on \textsc{RoboCasa}, and $15\%$ over OpenVLA~\citep{kim2024openvla} finetuned on \textsc{LIBERO}.
On \textsc{ReMemBench}, its performance scales with memory size: training with longer short-term memory windows (1$\xrightarrow{}$512 timesteps) yields progressively higher success rates (from $16\%$ to $42\%$) without saturation.
In a real-world adaptation of a \textsc{ReMemBench} task, PRISM improves absolute success by $15\%$ ($30$\% vs. $15$\%) over prior works, while a policy without memory achieves $0$\%.
Together, these results establish PRISM as an effective and scalable approach to visuomotor policies with short-term memory.

\definecolor{explicit}{HTML}{F2FAFF} %
\definecolor{implicitBase}{HTML}{FFEACC}
\colorlet{implicit}{implicitBase!100}  %

\begin{table*}[ht]
\centering
\setlength\extrarowheight{1mm}
\resizebox{0.95\textwidth}{!}{
\begin{tabular}{lcccccc}
\toprule
\multirow{2}{*}{\textsc{Benchmark}} & \multirow{2}{*}{\textsc{Domain}} & \multirow{2}{*}{\textsc{Episode Length}} & \multicolumn{4}{c}{\textsc{Short-term memory categories}} \\
\cline{4-7}
& & & \textsc{Spatial} & \textsc{Prospective} & \textsc{Object-Associative} & \textsc{Object-Set} \\
\midrule
\rowcolor{implicit} ALFRED~\citep{shridhar2020alfred} & Manipulation & 1000 & \checkmark & \xmark & \xmark & \xmark \\
\rowcolor{implicit} Habitat MultiON~\citep{wani2020multion} & Navigation & 2500 & \checkmark & \xmark & \xmark & \xmark \\
\rowcolor{implicit} FindingDory~\citep{yadav2025findingdory} & Navigation + manipulation & 3500& \checkmark & \checkmark & \xmark & \xmark \\
\rowcolor{explicit} Memory Maze~\citep{pasukonis2022evaluating} & Navigation & 4000 & \checkmark & \xmark & \xmark & \xmark \\
\rowcolor{explicit} Memory Gym~\citep{pleines2023memory} & Synthetic gridworlds & -1 & \checkmark & \xmark & \checkmark & \xmark \\
\rowcolor{explicit} POPGym~\citep{morad2023popgym} & Synthetic games & 1024 & \checkmark & \xmark & \checkmark & \checkmark \\
\rowcolor{explicit} MemoryBench~\citep{fang2025sam2act} & Manipulation & 12 & \checkmark & \xmark & \checkmark & \xmark \\
\rowcolor{explicit} Mikasa-Robo~\citep{cherepanov2025memorybenchmarkrobots} & Manipulation & 180 & \checkmark & \checkmark & \xmark & \xmark \\
\rowcolor{explicit} RoboMME~\citep{robomme} & Manipulation & 1300 & \checkmark & \xmark & \checkmark & \checkmark \\
\midrule
\rowcolor{explicit} \textbf{\textsc{ReMemBench}} & Navigation + manipulation & 2500 & \checkmark & \checkmark & \checkmark & \checkmark \\
\bottomrule
\end{tabular}
}
\vspace{-4pt}
\caption{Comparison of memory-focused embodied AI benchmarks. Rows highlighted in \colorbox{explicit}{blue} explicitly require memory by design (i.e., identical observations can correspond to different actions depending on history). Rows in \colorbox{implicit}{orange} involve memory implicitly but do not explicitly test for it. \textsc{ReMemBench} uniquely covers all short-term categories in visuomotor tasks.}
\label{tab:memory_benchmarks}
\end{table*}

\section{Related Work}
\paragraph{Short-Term Memory in Machine Learning}
Recurrent architectures such as RNNs~\citep{elman1990finding}, LSTMs~\citep{hochreiter1997long}, and more recent state-space models like Mamba~\citep{gu2023mamba} have been widely used to capture temporal dependencies. Their memory capacity, however, is tied to the parameter count, and information is only indirectly accessible via hidden states, leading to the credit assignment problem in backpropagation through time~\citep{bengio1993credit}.
Transformers address these limitations by decoupling memory capacity from parameters and enabling direct access to stored representations~\citep{vaswani2017attention}. Yet, they introduce new challenges: attention scales linearly with context length at test time, visual inputs yield high token costs (often $10^2$ per image), and irrelevant tokens can degrade performance~\citep{hong2025context}.
\rebuttal{
Stability and efficiency have been pursued through attention sinks~\citep{xiao2023efficient}, gated attention~\citep{qiu2025gated}, hierarchical attention~\cite{yang2016hierarchical}, hybrid attention mechanism with global and sliding window~\cite{beltagy2020longformer}, attention with segment-level recurrence~\cite {dai2019transformerxlattentivelanguagemodels,parisotto2019stabilizingtransformersreinforcementlearning}, but these remain largely unexplored in closed-loop visuomotor control.
}

\paragraph{Short-Term Memory in Visuomotor Policies} 
In visuomotor policies, short-term memory has traditionally been handled by recurrent policies such as LSTMs~\cite{payattention,mandlekar2021matterslearningofflinehuman}
In practice, however, these models often struggle to scale to long-horizon manipulation, where relevant information may be separated by hundreds of timesteps. 
Recent transformer-based visuomotor policies trained via imitation learning largely discard history, conditioning only on the current frame or a short window~\citep{chi2023diffusion, zheng2024tracevla, black2024pi0, bjorck2025gr00t, team2025gemini}.
Conversely, some methods aggressively compress entire interactions into a few embedding vectors to reduce computation, discarding fine-grained temporal and spatial information~\cite{fang2019scenememorytransformerembodied,shi2025memoryvla}.
To address this limitation, recent work incorporates specialized memory representations into transformer-based policies, such as the spatial memory in SAM2Act++~\citep{fang2025sam2act} or memory constructed by storing selected keyframes using hand-designed rules~\cite{memer}.
Auxiliary objectives provide another route, as in Past-Token Prediction (PTP)~\citep{torne2025learning}, which reconstructs past tokens to regularize policies.
In contrast, our work explores an alternative: improve memory utilization of the transformer backbone through gated attention and hierarchical architecture, enabling selective retention of information and efficient scaling without external modules or auxiliary losses.

\paragraph{Memory Benchmarks} 
Several embodied benchmarks involve partial observability, where memory is useful but not isolated. ALFRED~\citep{shridhar2020alfred}, Habitat MultiON~\citep{wani2020multion}, and FindingDory~\citep{yadav2025findingdory} ask agents to recall object states or past goals, but performance also depends heavily on perception, semantics, and exploration. In contrast, recent work introduces benchmarks that explicitly enforce memory dependence. Memory Maze~\citep{pasukonis2022evaluating} evaluates spatial recall in procedurally generated mazes. Memory Gym~\citep{pleines2023memory} and POPGym~\citep{morad2023popgym} provide gridworld and game-based tasks targeting short-term recall.
MemoryBench~\citep{fang2025sam2act} defines manipulation tasks with identical observations yield different actions depending on history, but mainly targets a single spatial memory type where specialized methods already reach near-saturated performance.
Mikasa-Robo~\citep{cherepanov2025memorybenchmarkrobots} instead studies learning algorithms with memory for relatively short-horizon manipulation (episodes up to $\sim$180 steps). In contrast, ReMemBench, with its human teleoperation dataset, is designed to study imitation learning under partial observability. It spans both navigation and manipulation with longer-horizon, multi-subtask episodes (up to $\sim$2.5k steps), organized into four memory categories: spatial, object-set, object-associative, and prospective.
More recently, RoboMME~\cite{robomme} offers a manipulation benchmark with diverse tasks and extended horizons, but lacks coverage of prospective memory tasks while including additional tasks involving procedural memory. ReMemBench complements this landscape by targeting four different memory categories across both navigation and manipulation.

In summary (\autoref{tab:memory_benchmarks}), \textsc{ReMemBench} encompasses short-term memory of various categories, encouraging unified memory architectures that generalize across tasks.

\section{\textbf{PRISM}: A \textbf{\underline{P}}olicy A\textbf{\underline{r}}ch\textbf{\underline{i}}tecture for \textbf{\underline{S}}hort-Term \textbf{\underline{M}}emory }

\begin{figure*}[t]
\begin{center}
    \includegraphics[width=0.95\textwidth]{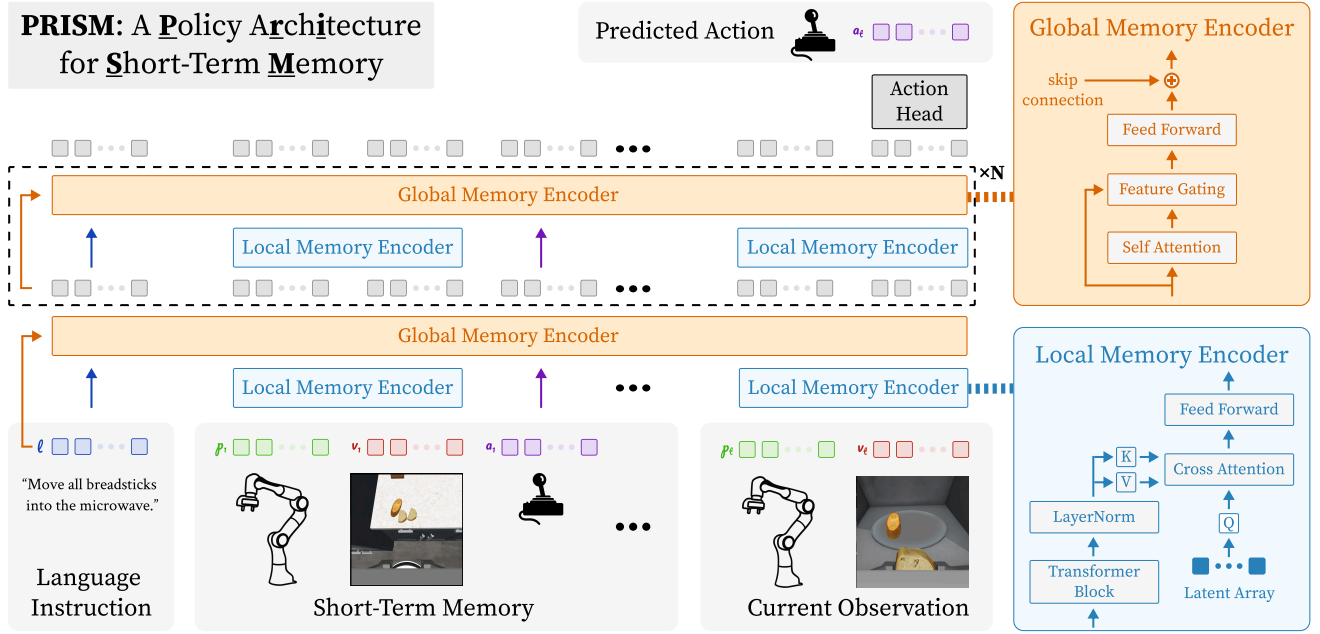}
    \vspace{-0.1cm}
    \caption{PRISM applies gated attention to filter information retrieved from history and hierarchical summarization to scale attention over long interaction histories, improving causal transformer policies trained with behavior cloning by improving robustness to noisy histories and reducing computation.}
    \vspace{-0.6cm}
    \label{fig:architecture}
\end{center}
\end{figure*}

\subsection{Formulation, Background, and Challenges}
\paragraph{Problem Formulation}
We study imitation learning under partial observability, where the robot receives observations $o_t$ (\textit{e.g.}, RGB images and proprioception) and produces actions $a_t$ in a closed loop. The training dataset $\mathcal{D}_{\text{train}}=\{(\tau^i, l^i)\}_{i=1}^{N}$ consists of $N$ expert trajectories $\tau^i=\{(o^i_t,a^i_t)\}_{t=1}^{T^i}$, paired with a task language instruction $l^i$.  
We learn a policy $\pi_\theta$ via behavior cloning, conditioning on the interaction history up to time $t$:
\[
a_t \sim \pi_\theta\!\left(a \mid o_{\le t}, a_{<t}, l\right).
\]

\paragraph{Training Objective}
We optimize a behavior cloning loss to match the expert distribution,
\[
\mathcal{L}(\theta)
= - \mathop{\mathbb{E}}\limits_{(\tau,l)\sim \mathcal{D}_{\text{train}}}
    \sum_{t=1}^{T(\tau)}
    \log p_\theta\!\left(a_t \mid o_{\le t}, a_{<t}, l\right).
\]
where $p_\theta(\cdot)$ denotes a parametric action distribution.  
We employ action chunking, predicting the next $l$ future actions at each timestep (not shown above for clarity)~\citep{zhao2023learning}.

\paragraph{Evaluation}
During inference, the policy predicts actions autoregressively with language tokens forming a static prefix, while observation and action tokens accumulate over time. Once the horizon budget $n$ is reached, only the most recent $n$ steps are retained in memory. The learned policy is evaluated on its ability to generalize to new object placements.

\paragraph{Model Architecture}
At each timestep $t$, the policy processes a truncated history
\[
H_t = (o_k,a_k,o_{k+1},a_{k+1},\ldots,o_{t-1},a_{t-1},o_t),
\]
together with the task description $l$, where $k = \max(t-n, 1)$, $n$ is a fixed memory budget on the number of retained interaction steps. Each observation $o_t = (v_t, p_t)$ consists of an image $v_t$ and a proprioceptive state $p_t$. Images are encoded into $k_v$ visual tokens using an image encoder, proprioceptive states into $k_p$ tokens, and past actions into $k_a$ tokens via an action encoder. The instruction $l$ is encoded with a pretrained text encoder into $k_l$ language tokens.  

All tokens are packed in temporal order with a causal mask, and positional encodings preserve temporal and (for images) spatial structure. The resulting token sequence is
\begin{equation}
\begin{aligned}
[l]_{1:k_l}, [p_k]_{1:k_p}, [v_k]_{1:k_v}, [a_k]_{1:k_a},\ldots, [p_t]_{1:k_p}, [v_t]_{1:k_v}.
\end{aligned}
\end{equation}
This sequence is processed by a transformer-based backbone~\citep{vaswani2017attention}. The transformer maintains the sequence within its context window, which serves as the short-term memory. At each step, the hidden state of the last observation token $[v_t]_{k_v}$ is passed through an action head to produce the parameters of $p_\theta(a_t \mid \cdot)$.

\paragraph{Challenges}
Attending to long histories is crucial for tasks that require memory but raises two key challenges.
First, long histories often include irrelevant information that induces spurious correlations during training.
Second, naively attending to all past tokens leads to prohibitive memory and computation costs during both training and evaluation.

\begin{figure*}[ht]
\begin{center}
    \includegraphics[width=0.98\textwidth]{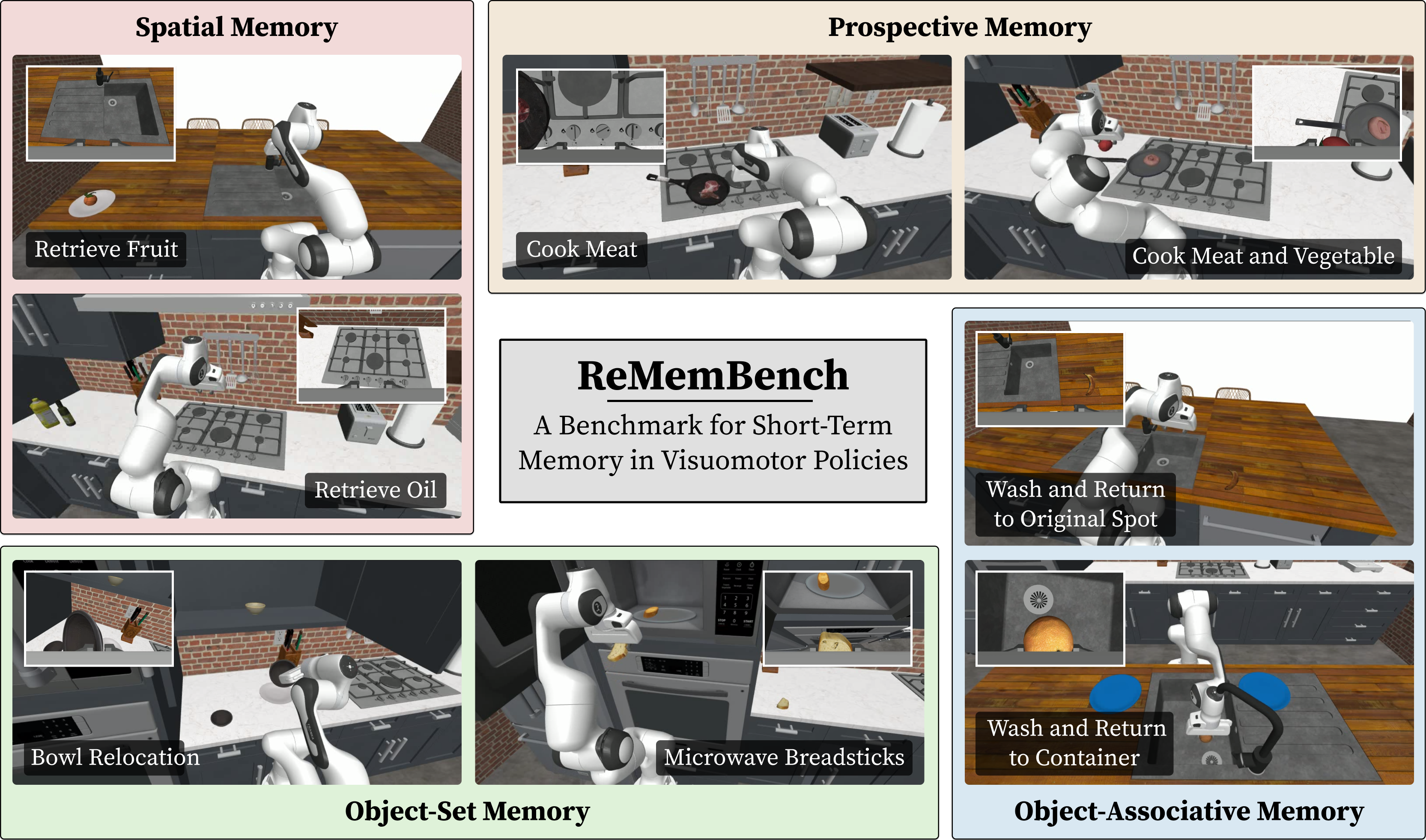}
    \vspace{-0.2cm}
    \caption{Guided by the cognitive science literature, ReMemBench decomposes short-term memory into several functional categories, such as Spatial, Prospective, Object-Set, and Object-Associative.}
    \vspace{-0.6cm}
    \label{fig:benchmark}
\end{center}
\end{figure*}

\subsection{Selecting Relevant Features}
Transformers retain a history of inputs in their context window, effectively forming the short-term memory of a policy.
However, long input histories can also introduce irrelevant information, leading models to learn spurious correlations~\citep{de2019causal,wen2020fighting}, \textit{e.g.}, associating the positions of past distractor objects with a decision to place an object on the left or right.
At test time, such correlations may not hold true, resulting in poor performance. This issue is especially pronounced in behavior cloning, where even a small initial error $\epsilon$ can compound over time as $O(T^2\epsilon)$ for horizon length $T$~\citep{ross2011reduction, spencer2021feedback}.
Thus, while memory is essential for long-horizon household tasks, it also introduces noise that makes learning more difficult.

Transformers with longer context windows are particularly vulnerable to this issue due to the nature of the attention mechanism: the softmax operation distributes normalized weights throughout the history, potentially amplifying the influence of irrelevant tokens~\citep{peters2019sparse,hong2025context}.

To mitigate this, we use a gating mechanism that regulates the information flow after the attention~\citep{qiu2025gated}. Specifically, we modulate the output of each attention block using a learnable gate conditioned on the current input features.
Conditioning the gate on the input features allows the model to adaptively control which retrieved information to suppress as the interaction progresses.

Let $x$ denote the input to a transformer block, and let
\[
z = \text{Softmax}\!\left(\frac{QK^\top}{\sqrt{d}}\right)V
\]
denote the attention output after the softmax-weighted value aggregation but before the output projection $W_o$. We compute the gated output as:
\[
\tilde{z} = g(x) \odot z,
\]
where $g : \mathbb{R}^{d} \to (0,1)^d$ is a gating function, implemented as an MLP with sigmoid activation, and $\odot$ denotes element-wise multiplication. The gated output is first projected using the attention output projection $W_o$, and then incorporated via the standard residual connection: $y = x + W_o \tilde{z}$, followed by the feed-forward layer (See~\autoref{fig:architecture}, top-right).

\subsection{Reducing Memory and Compute Footprint}
Naively attending over all tokens from $n$ timesteps in history, each with $k_v$ visual tokens, $k_p$ proprioceptive tokens, and $k_a$ action tokens, incurs a quadratic compute and memory cost during training:
\[
\mathcal{O}\!\big((n k_{\text{tot}})^2\big)\text{, where } k_{\text{tot}} = k_p + k_v + k_a
\]
Similarly, during evaluation, it incurs a linear cost per step, $\mathcal{O}(n k_{\text{tot}})$, assuming key-value caching.
However, as the context grows, the compute and memory costs can become prohibitive.

To improve scalability, we propose a hierarchical architecture that first splits the sequence into a local encoding of memory, then aggregates the compressed outputs using a global memory encoder.
This approach avoids processing the entire token set at once and splits the computation into two parts~\citep{arnab2021vivitvideovisiontransformer}.
If each local window compresses $k_{\text{tot}}$ tokens down to $k_{\text{tot}} / m$, then the total compute and memory cost during training becomes: %
\[
O\big(nmk_\text{tot} + (nk_{\text{tot}}/m)^2\big)
\]
which is approximately $m^2\times$ cheaper than original $\mathcal{O}((n k_{\text{tot}})^2)$ when $m \ll nk_{\text{tot}}$.
Similarly, during evaluation, the hierarchical architecture yields a factor-$m$ gain in memory and compute: attention is computed over only $n k_{\text{tot}} / m$ tokens instead of $n k_{\text{tot}}$, and the number of cached key-value tokens is likewise reduced by $m$.

Concretely, at each step, a \emph{local memory encoder} first processes the current timestep's sensory information using a transformer block, then uses a perceiver-style resampler~\citep{jaegle2021perceiver} to compress $k_{\text{tot}} \!\to\! k_{\text{tot}}/m$ (\autoref{fig:architecture}, bottom-right).
A \emph{global memory encoder} then computes attention over the concatenated compressed tokens across time (\autoref{fig:architecture}, top-right).
The resulting token with global information is \emph{broadcasted} back to the original $k_{\text{tot}}$ tokens and fused with the original inputs via a residual skip to the local memory encoder.
The skip connection in PRISM helps preserve per-token individuality while injecting global information.

\paragraph{Summary.}
PRISM combines (i) gated attention, which adaptively filters irrelevant information from memory, and (ii) a hierarchical architecture, which reduces the compute and memory costs of incorporating long histories.
Together, they improve robustness to spurious correlations while allowing the policy to efficiently scale memory to longer horizons.
Both components integrate seamlessly with a transformer policy trained using a standard behavior cloning objective; no auxiliary losses or extra supervision are required.

\begin{table*}[t]
\vspace{-0.4cm}
\centering
\begin{minipage}[t]{0.7\linewidth}
\centering
\vspace{0pt}
\caption{Success rates on \textsc{ReMemBench} task categories (20 trials × 3 seeds). 
The four tasks are \textit{Retrieve Fruit}, \textit{Cook Meat}, \textit{Wash and Return to Container}, and \textit{Microwave Breadsticks} (see \autoref{sec:main:benchmarking}).}
\label{tab:results:main_table}
\vspace{-0.2cm}
\resizebox{\linewidth}{!}{%
\begin{tabular}{l l c c c c c}
\toprule
Group & Method & Spatial & Prospective & Obj-Assoc. & Obj-Set & Avg \\
\midrule
\multirow{2}{*}{\textsc{ReMemBench}}
  & LSTM~\citep{vennerod2021long}  & $0.14$ & $0.06$ & $0.15$ & $0.11$ & $0.12\pm0.04$ \\
  & Mamba~\citep{gu2023mamba} & $0.21$ & $0.52$ & $0.15$ & $0.14$ & $0.25\pm0.03$ \\
\multirow{7}{*}{(Four Tasks)}
  & TrXL~\citep{dai2019transformerxlattentivelanguagemodels} & $0.06$ & $0.28$ & $0.12$ & $0.12$ & $0.15 \pm 0.03$ \\
  & GTrXL~\citep{parisotto2019stabilizingtransformersreinforcementlearning} & $0.23$ & $0.38$ & $0.23$ & $0.21$ & $0.26\pm0.03$ \\
  & Linear-Attn.~\citep{katharopoulos2020transformersrnnsfastautoregressive} & $0.19$ & $0.30$ & $0.22$ & $0.16$ & $0.22\pm0.03$ \\
  & PTP~\citep{torne2025learning} & $0.28$ & $0.58$ & $0.16$ & $0.12$ & $0.28 \pm 0.05$ \\
  & SAM2Act++~\citep{fang2025sam2act} & $0.28$ & $0.08$ & $0.28$ & $0.24$ & $0.22\pm0.03$ \\
  & SMT~\citep{fang2019scenememorytransformerembodied} & $\hlc{0.41}$ & $0.40$ & $0.30$ & \hlc{$0.33$} & $0.36\pm0.02$ \\
  & PRISM & $0.28$ & \hlc{$0.77$} & \hlc{$0.33$} & $0.25$ & \hlc{$0.41\pm0.04$} \\
\midrule
\midrule
\multirow{1}{*}{\textsc{ReMemBench (All)}}
  & PRISM & $0.39$ & $0.43$ & $0.22$ & $0.25$ & $0.32\pm0.08$ \\
\bottomrule
\end{tabular}
}
\end{minipage}
\hfill
\begin{minipage}[t]{0.29\linewidth}
\centering
\vspace{0pt}

\includegraphics[width=\linewidth,height=0.45\textheight,keepaspectratio]{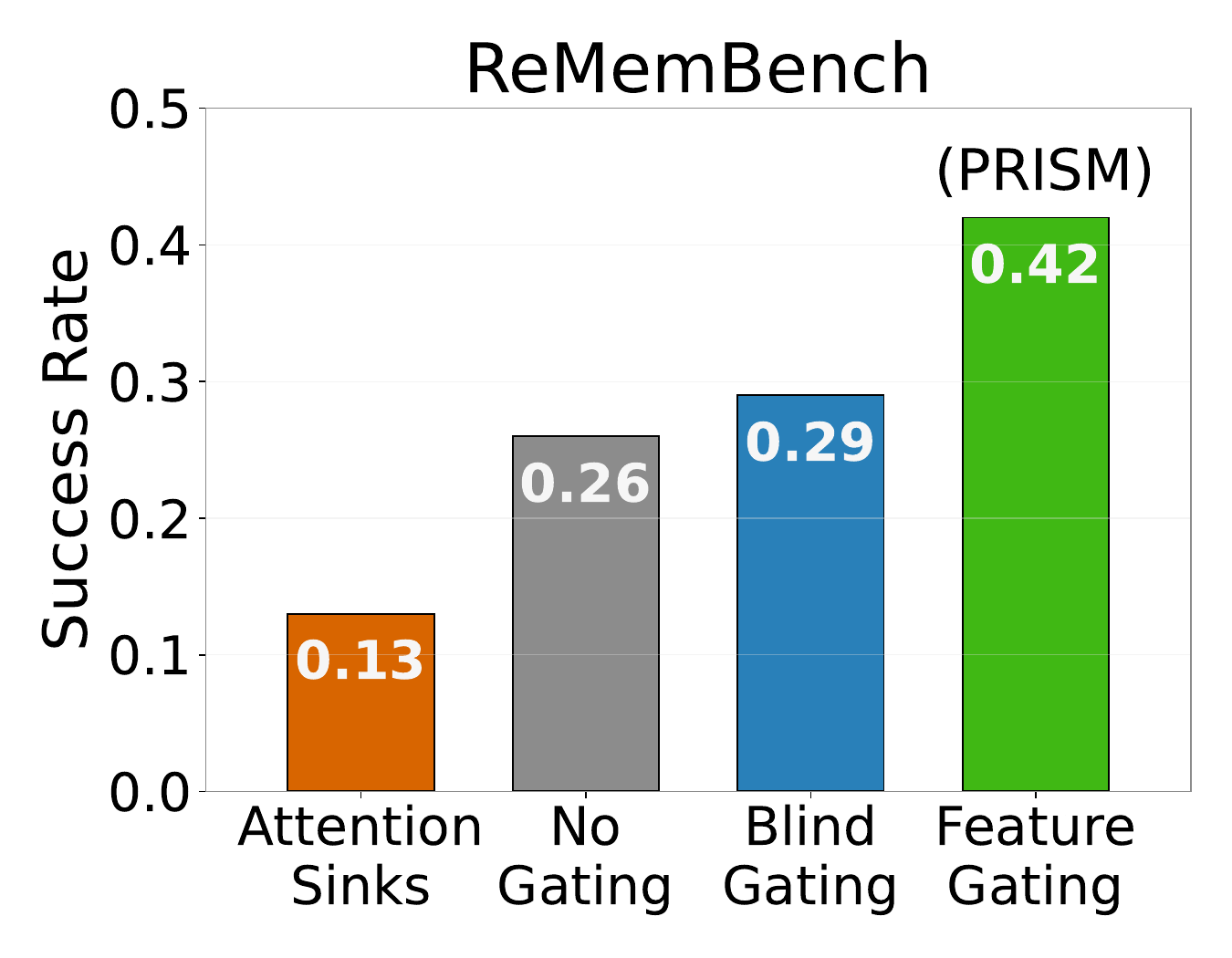}

\vspace{-6pt}
\captionsetup{type=figure}
\caption{Comparison with alternative mechanisms to the proposed feature-wise gating in PRISM.}
\label{fig:results:gating_variants}
\end{minipage}
\vspace{-10pt}
\end{table*}

\section{ReMemBench: A Benchmark for Short-Term Memory in Visuomotor Policies}\label{sec:main:benchmarking}
\textsc{ReMemBench} is designed to evaluate short-term memory in visuomotor policies. 
Guided by the cognitive science literature, we decompose short-term memory into several functional categories (\autoref{fig:benchmark}).
Diversity across categories promotes general memory mechanisms rather than task-specific, non-generalizable solutions.
Each category is instantiated with two household manipulation tasks.
All tasks ensure that immediate sensory inputs are insufficient and success requires within-episode recall.\\
\noindent
\textbf{Spatial Memory.} 
The ability to recall object locations~\citep{o1978hippocampus}. We created the following tasks, taking inspiration from the ``Belongings: Delayed Recall'' subtest of the Rivermead Behavioral Memory Test~\citep{wilson1985rbmt}: 
\begin{itemize}[leftmargin=1em, noitemsep, topsep=0pt]
\item \emph{Retrieve Fruit:} A fruit is placed somewhere in the kitchen and is out of view from a neutral start. The robot must remember its location, fetch it, and place it in the sink. \rairebuttal{Fruit type and object placement differ across trials.}
\item \emph{Retrieve Oil:} Several household items are arranged and not visible from the start. The robot must find the specified oil bottle and pick it up. The bottle instance and placement vary across trials.
\end{itemize}
\noindent
\textbf{Prospective Memory.} 
The ability to retain intentions over a delay and execute them at the right time~\citep{einstein1990normal}. We consider the following tasks, inspired by the Prospective-Retrospective Memory Questionnaire~\citep{smith2000prospective}:
\begin{itemize}[leftmargin=1em, noitemsep, topsep=0pt]
\item \emph{Cook Meat:} A pan with meat sits on the stove. The robot turns the stove on, waits the required duration, then turns it off. Target duration, pan, and meat type vary across trials.
\item \emph{Cook Meat and Vegetable:} Meat begins on the stove, and a vegetable is nearby. The robot must turn the stove on, add the vegetable after a specified duration, and turn the stove off after another specified duration so both items meet their cooking times. Durations, pan type, ingredient types, and vegetable location vary across trials.
\end{itemize}
\noindent
\textbf{Object-Associative Memory.} 
The ability to recall object-location associations~\citep{mayes2007associative}. 
We propose the following, inspired by the Paired Associates Learning test~\citep{sahakian1988comparative}: 
\begin{itemize}[leftmargin=1em, noitemsep, topsep=0pt]
\item \emph{Wash and Return to Container:}
Two saucers sit to the left and right of the sink; one holds a fruit. The robot must wash the fruit and return it to the \emph{same} saucer. Fruit and saucer types, and saucer sides vary across trials.
\item \emph{Wash and Return to Original Spot:}
A fruit starts on the countertop. The robot moves it to the sink to wash it, then returns it to its original spot (within a small window). Initial positions and fruit type vary across trials.
\end{itemize}
\noindent
\textbf{Object-Set Memory.} 
The ability to maintain and update sets of multiple objects across time~\citep{luck1997capacity,makovski2009role}. We consider the following tasks, inspired by object set maintenance paradigms in working memory, particularly the Counting Span task~\citep{case1982operational}:
\begin{itemize}[leftmargin=1em, noitemsep, topsep=0pt]
\item \emph{Microwave Breadsticks:} A plate holds multiple breadsticks, and the microwave is initially out of view. The robot must move all breadsticks into the microwave, close the door, and keep track of how many remain. Counts, bread types, and object positions vary across trials.
\item \emph{Relocate Bowls:} Bowls with distractor plates nearby sit beside a cabinet. The robot must transfer all and only the bowls into the cabinet while tracking the remaining count. Bowl types, counts, and object positions vary across trials.
\end{itemize}

\textsc{ReMemBench} builds on assets from the \textsc{RoboCasa} environment~\citep{nasiriany2024robocasa}, with each task provided with $50$ expert teleoperated demonstrations for training.

\vspace{-0.1cm}
\section{Experiments}\label{sec:experiments}
\textbf{Baselines.}
We group our baselines into four categories according to their memory mechanisms and application domains and evaluate them on \textsc{ReMemBench}.
\textbf{Recurrent baselines} include LSTM~\citep{vennerod2021long} and Mamba~\citep{gu2023mamba}, which compress history into hidden states.
\textbf{Segment-level recurrent transformers} include Transformer-XL (TrXL)~\citep{dai2019transformerxlattentivelanguagemodels} and Gated Transformer-XL (GTrXL)~\citep{parisotto2019stabilizingtransformersreinforcementlearning}, which combine recurrence with attention.
We also include \textbf{transformer-variants} such as linear-attention transformer~\citep{katharopoulos2020transformersrnnsfastautoregressive}, which replaces softmax with a linear kernel to obtain long-context attention with reduced compute and memory.
Finally, under \textbf{visuomotor policy architectures} for imitation under partial observability, we consider Past Token Prediction (PTP)~\citep{torne2025learning}, which adds an auxiliary loss to predict past actions, Scene Memory Transformer (SMT)~\citep{fang2019scenememorytransformerembodied}, which compresses entire trajectories into single token via attention pooling, and SAM2Act++~\citep{fang2025sam2act} that creates a compact state representation by storing the pixel co-ordinates of objects observed in the past along with the current observation.

Our evaluations answer the following research questions.

\textbf{Q1. How does PRISM compare to prior works on imitation learning with partial observability?}
\textit{\textsc{ReMemBench} Results.}
Recurrent models such as LSTM and Mamba compress the entire history into a limited capacity hidden state and struggles with assigning credits to the correct timestep using backpropagation through time, \textit{e.g.}, in a ``turn off the stove after a delay'' task, the learning signal from the decision should backpropagate through hundreds--thousands of updates, so gradients vanish~\citep{bengio1993credit}. Consequently, LSTM and Mamba achieve only $12\%$ and $25\%$ success versus PRISM's $41\%$ (absolute drop of $29$ and $16$ points, resp.) (Table \ref{tab:results:main_table}).
Segment-level recurrent transformers like TrXL and GTrXL still rely on recurrence, so long-horizon credit assignment remains challenging, yielding $15\%$ and $26\%$ (absolute drop of $26$ and $15$ points, resp.).
Linear-Attention effectively compresses the entire history into a single matrix; this representation likely suffers from forgetting over long horizons and reaches only $22\%$ (an absolute $19$-point drop)~\citep{he2025alleviatingforgetfulnesslinearattention}.

Prior standard attention-based methods also struggle to regulate the retrieved information.
PTP improves recall via an auxiliary loss but does not filter irrelevant information, leading to only $28\%$ success ($-14$ points).
Scene-level memory models like SMT compress entire trajectories into a pooled embedding, discarding fine-grained temporal information, reaching $36\%$ ($-5$ points).
In contrast, PRISM preserves token-level features and combines attention with feature-wise gating to retrieve information while suppressing irrelevant information.
Lastly, SAM2Act++ stores object spatial coordinates in memory. However, due to the lack of semantic information, this variant achieves only $22\%$ success ($-19$ points) in \textsc{ReMemBench}. This gap underscores the need to develop a general mechanism and architecture for short-term memory in visuomotor policies, rather than relying on task-specific design choices.

\textit{Real-World Results.} 
\begin{figure}[h]
    \centering
    \vspace{-5pt}
    \includegraphics[width=\linewidth]{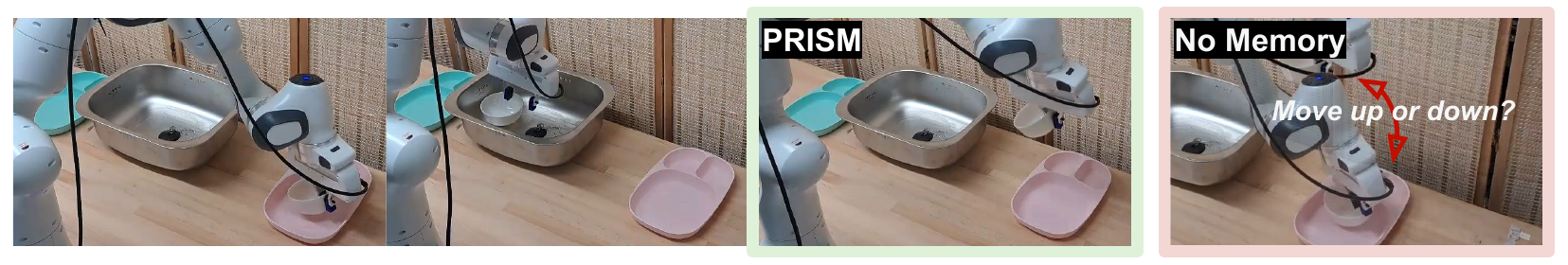}
    \vspace{-20pt}
    \caption{(Left) Successful real-world rollout of PRISM on \emph{Wash and Return to Container}. (Right) Without memory, the policy cannot disambiguate `just picked up' vs. `about to place.'}
    \label{fig:real_robot}
\end{figure}

We evaluate PRISM on a real-world adaptation of the `Wash and Return to Container' task from \textsc{ReMemBench} (\autoref{fig:real_robot}).
Without access to memory, the baseline policy fails completely ($0\%$).  
The transformer-based PTP baseline achieves a $15\%$ success, while PRISM reaches $30\%$.  
This substantial improvement demonstrates PRISM's transfer to real-world tasks, where temporally extended short-term memory is important.

\begin{table*}[t]
\centering
\footnotesize
\setlength{\tabcolsep}{3pt}

\begin{minipage}[t]{0.30\textwidth}
\centering
\resizebox{\linewidth}{!}{%
\begin{tabular}{lc}
\toprule
Method & Success Rate \\
\midrule
GR00T-N1-2B       & 0.32 \\
Diffusion Policy  & 0.26 \\
PRISM (no memory) & 0.32 \\
PRISM ($n=256$)   & \textbf{0.43} \\
\bottomrule
\end{tabular}}
\label{tab:robocasa_results}
\end{minipage}\hspace{0.02\textwidth}%
\begin{minipage}[t]{0.64\textwidth}
\centering
\resizebox{\linewidth}{!}{%
\begin{tabular}{lcccccc}
\toprule
Method & LIBERO-90 & LIBERO-10 & Object & Spatial & Goal & Avg. \\
\midrule
OpenVLA           & 0.62          & 0.54 & 0.88 & 0.85 & 0.79 & 0.74 \\
Diffusion Policy  & --            & 0.72 & \textbf{0.93} & 0.78 & 0.68 & -- \\
PRISM (no memory) & 0.77          & 0.75 & 0.89 & 0.85 & 0.61 & 0.77 \\
PRISM ($n=256$)   & \textbf{0.85} & \textbf{0.81} & \textbf{0.93} & \textbf{0.92} & \textbf{0.95} & \textbf{0.89} \\
\bottomrule
\end{tabular}}
\end{minipage}

\caption{Success rates on \textsc{RoboCasa} (Left) and \textsc{LIBERO} (Right). PRISM with memory ($n=256$) outperforms its no-memory variant and strong pretrained baselines finetuned on \textsc{RoboCasa} and LIBERO. Baseline numbers are adopted from prior works~\citep{bjorck2025gr00t, kim2024openvla, belkhale2024minivla}.}
\label{tab:robocasa_libero_results}
\end{table*}

\textbf{Q2. How much improvement does PRISM provide on standard benchmarks that do not explicitly test memory?}
Augmenting the state with recent observations provides information absent from the current observation.
This disambiguates visually identical inputs that require different actions (\textit{e.g.}, ``just grasped'' vs. ``about to place''), reducing multimodality in the state-conditioned action distribution.

Empirically, we evaluate PRISM on two visuomotor benchmarks, \textsc{RoboCasa} and LIBERO. On \textsc{RoboCasa}, PRISM with memory improves over its no-memory variant by an absolute $11$ points ($32\%\rightarrow43\%$) and also exceeds GR00T-N1-2B finetuned on \textsc{RoboCasa} by $11$ points (\autoref{tab:robocasa_libero_results}).
On LIBERO, PRISM improves similarly over its no-memory counterpart by $12$ points, averaged over all benchmarks, and pretrained OpenVLA finetuned on LIBERO by $15$ points (\autoref{tab:robocasa_libero_results}).
These results indicate that adding short-term memory using PRISM improves both memory-dependent and seemingly memory-less tasks by adding information, making the conditional action distribution less multimodal.

\begin{figure*}[h]
    \vspace{-10pt}
    \centering
    \begin{minipage}[t]{0.7\textwidth}
        \centering
        \includegraphics[width=\linewidth]{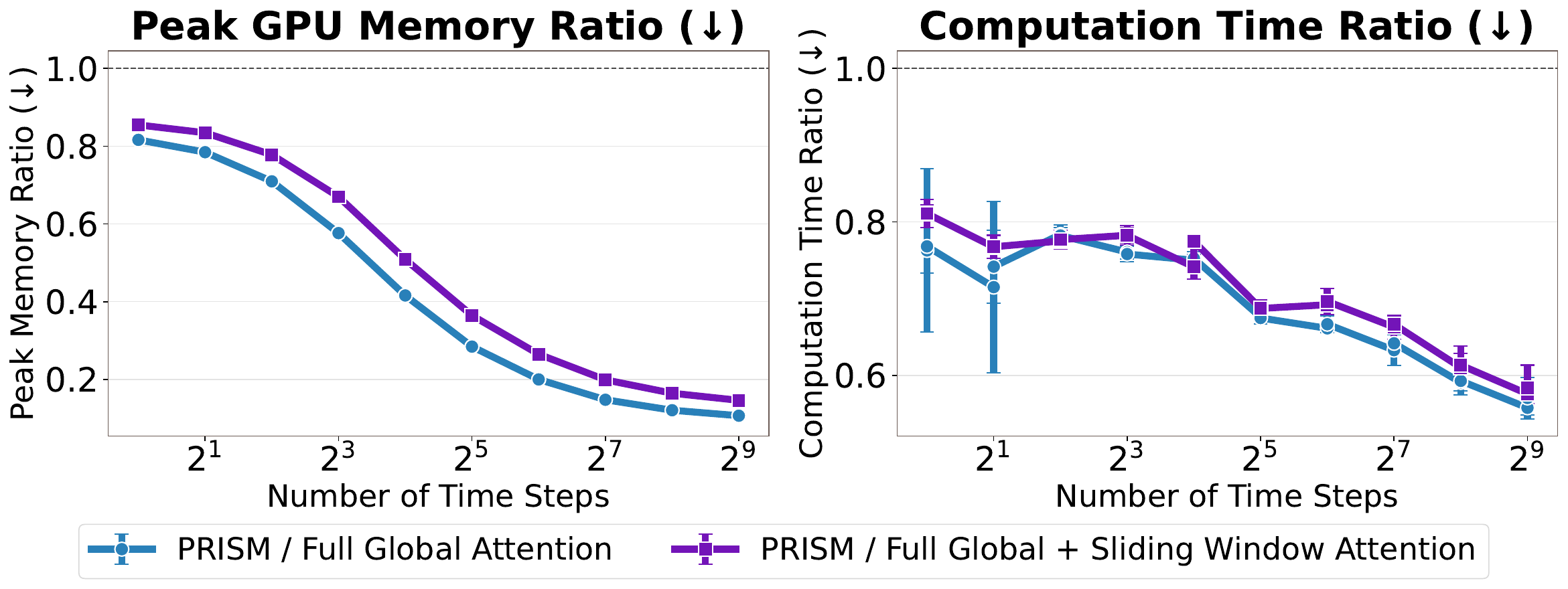}
    \end{minipage}\hfill
    \begin{minipage}[t]{0.28\textwidth}
        \centering
        \includegraphics[width=\linewidth, trim=0 0 0 0cm]{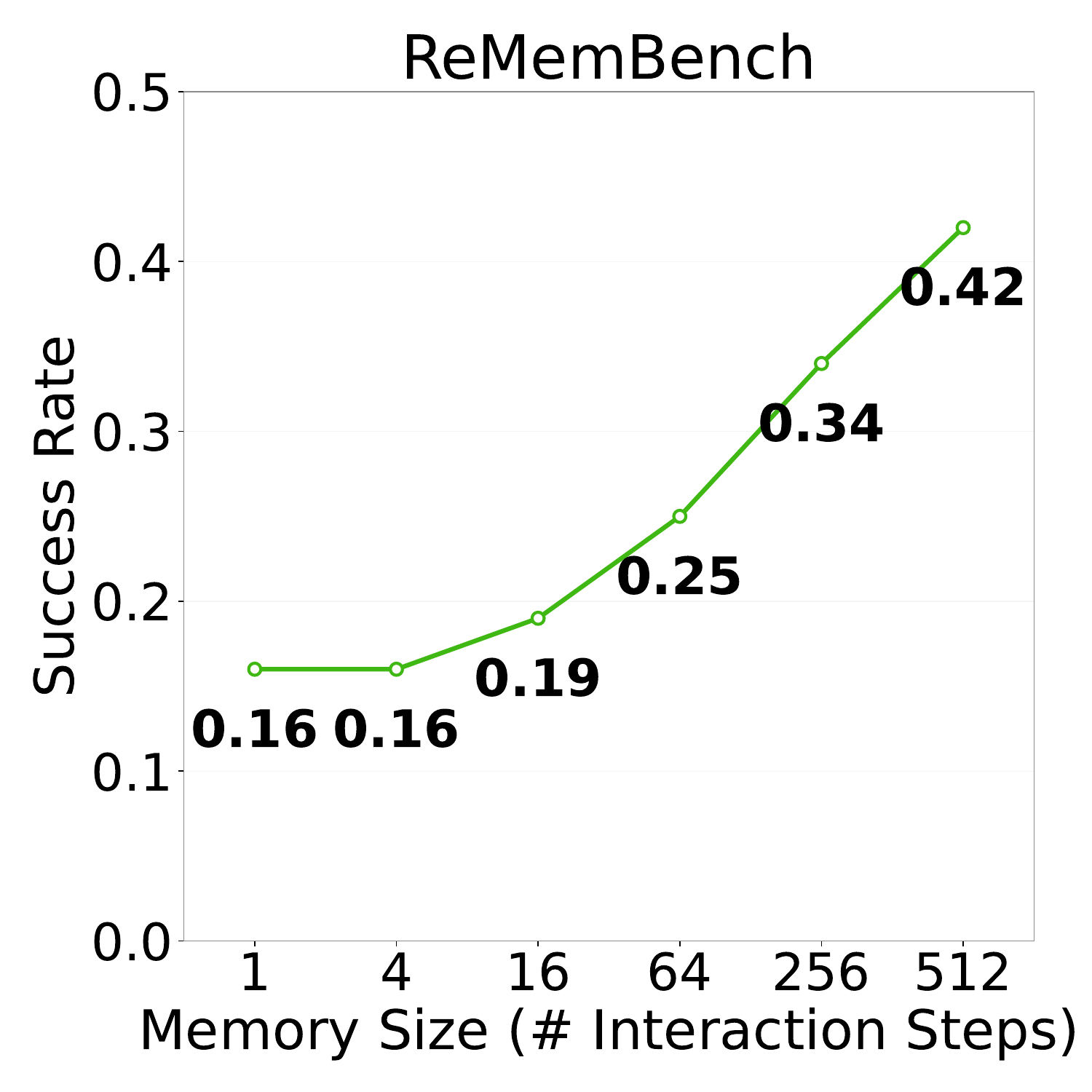}
    \end{minipage}
    \vspace{-1.5pt}
    \caption{(Left) PRISM is both computationally and memory efficient at handling long contexts. (Right) PRISM demonstrates strong scalability as memory size increases on \textsc{ReMemBench} tasks.}
    \label{fig:results:increasing_memory}
    \label{fig:results:compute_benefits}
\end{figure*}

\textbf{Q3. How does PRISM’s performance scale with increasing memory capacity?}
PRISM exhibits strong scalability with increasing memory size on \textsc{ReMemBench} (\autoref{fig:results:increasing_memory}, Right).
As the memory window expands from $n=1$ to $n=512$, the success rate increases by $0.26$, indicating that PRISM continues to extract useful information from temporally extended memory without saturation.
This trend highlights PRISM's suitability for tasks that demand reasoning over extended temporal information.

\begin{figure}
    \centering
    \includegraphics[width=0.95\linewidth]{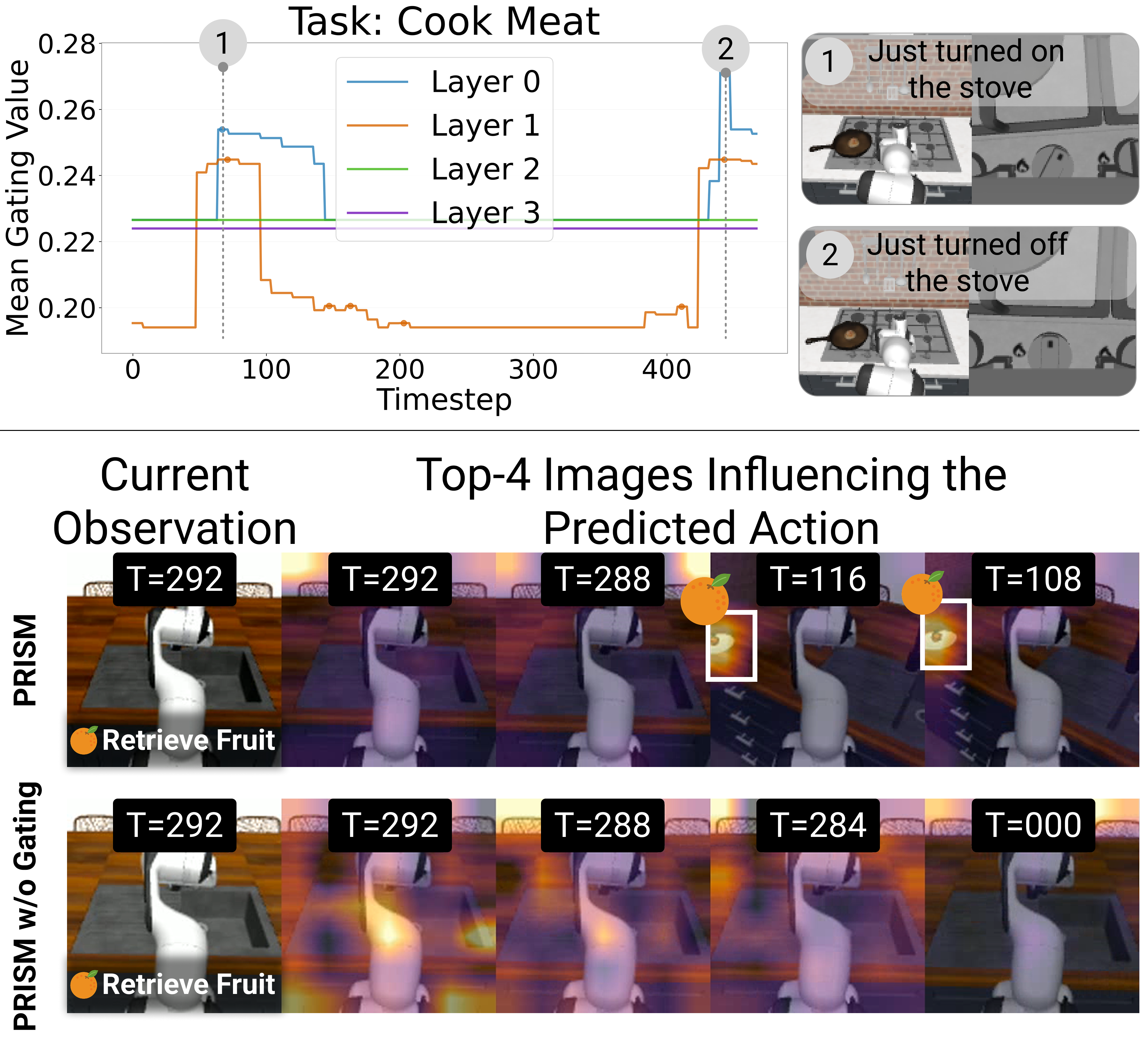}
    \vspace{-2pt}
    \caption{ \textbf{(\textit{Top}) Mean predicted gating value across timesteps during decision-making.} Higher values align with intuitive task moments, especially when turning the stove on and off, where memory recall is most important for disambiguation. This trend is consistent across successful ``Cook Meat'' trials. 
    \textbf{(\textit{Bottom}) Qualitative visualization of the gating effect.} With gating, the policy pays attention to the correct observations and the spatial region where the orange was previously observed from memory. Without gating, attention shifts to task-irrelevant regions, such as the sink.}
    \label{fig:results:qualitative}
\end{figure}

\textbf{Q4. Is gated attention effective for filtering irrelevant information in visuomotor policies?}
The gating mechanism in PRISM is crucial for selectively controlling information retrieved from memory; removing it leads to a $16$-point performance drop, underscoring the need to filter irrelevant information (\autoref{fig:results:gating_variants}, right).
We also evaluate a variant that uses a per-layer scalar gate while ignoring per-step input features. Although this offers a modest gain over no gating ($3$-points), it performs significantly worse than PRISM's input feature-aware gating ($29\%$ vs. $42\%$).
Furthermore, we test adding attention sink tokens instead of gating, which allows the model to attend to a dummy token rather than irrelevant information from memory~\citep{xiao2023efficient}.
While successful in language models~\citep{openai2025gptoss}, attention sinks prove ineffective, highlighting the need for further investigation into different memory filtering mechanisms in visuomotor policies. See~\autoref{fig:results:qualitative} (bottom) for a qualitative example.

\textbf{Q5. How much computational and memory efficiency is gained from the hierarchical architecture?}
PRISM scales efficiently over long temporal sequences by compressing local information before global attention, allowing global operations to run on a compact set of summary tokens rather than the full sequence. Moreover, it only needs to cache the summary tokens, rather than all the tokens. At $n=512$, this yields $0.20\times$ peak GPU memory ($-80\%$) and $0.60\times$ runtime ($-40\%$) relative to both full attention and hybrids with global attention + sliding window attention~\citep{beltagy2020longformer} (\autoref{fig:results:compute_benefits}, left). 
We observe an increase in the hybrid method's runtime performance at $n=4$, likely due to CUDA kernel scheduling rather than algorithmic effects. %

\section{Limitations and Conclusion}
\noindent
\textbf{Benchmark headroom.} Absolute success rates on \textsc{ReMemBench} cluster around the $30\%$ range. 
We view this not as a weakness of memory, but as evidence that long-horizon tasks are challenging. Improving overall performance is essential to demonstrate the benefits of history and to encourage improvements across components (\textit{e.g.}, perception, control, memory), rather than any single module.
\textbf{Short vs.\ long-term memory.} In this work, we focus on \emph{short-term} memory: retaining a recent window of sensory–motor history (on the order of minutes) for action prediction~\cite{atkinson1968human}. Looking ahead, we envision adding a persistent \emph{long-term} memory that accumulates information over days, or the system's lifetime and retrieves relevant traces into short-term memory when needed for decision-making.
\textbf{Internet-scale priors.} Learning to \emph{use} short-term memory effectively can require large amounts of data. Our study operates at a much smaller scale. An exciting direction is to pretrain memory modules on internet-scale video and study how these priors transfer to visuomotor control.

In conclusion, we introduce \textsc{PRISM}, a visuomotor policy architecture that incorporates short-term memory through gated attention and a hierarchical design. To systematically evaluate memory in visuomotor control, we propose \textsc{ReMemBench} with eight tasks spanning different categories: spatial, prospective, object-associative, and object-set.
PRISM outperforms state-of-the-art baselines across (1) all \textsc{ReMemBench} tasks, (2) \textsc{RoboCasa} atomic tasks not explicitly designed to test memory, and (3) real-world evaluations. Its performance also scales with memory capacity, with longer histories yielding consistent gains.
Together, PRISM and \textsc{ReMemBench} provide a foundation for developing and evaluating memory-augmented visuomotor policies for long-horizon tasks.

\clearpage
\section{Acknowledgements}
We thank Priya Sundaresan, Albert Yu, Zhenyu Jiang, Huihan Liu, Max Rudolph, Siddhant Agarwal, and anonymous reviewers at the Robotics and AI Institute (RAI) for their helpful feedback on the manuscript. We are also grateful to the Foundation Models and CAPTURE teams at RAI for providing resources and engaging discussions. We thank the teams at RAI and the Texas Advanced Computing Center (TACC) at The University of Texas at Austin for their invaluable support with computing resources.
This work was partially supported by the National Science Foundation (FRR-2145283, EFRI-2318065), the Office of Naval Research (N00014-22-1-2204, N00014-24-1-2550), and the DARPA TIAMAT program (HR0011-24-9-0428).
Finally, we note that this work was primarily conducted during the summer internships of Rutav Shah and Rajat Kumar Jenamani at Robotics and AI Institute.
\bibliography{references/psychology, references/methods, references/benchmarks}

\clearpage
\section{Appendix}
\section{Effect of Attention Scale Factor (Sharpness)}
\begin{table*}[ht!]
\centering
\small
\setlength{\tabcolsep}{5pt}
\begin{tabular}{lccccccc}
\toprule
$V=\text{scale\_factor} \cdot \sqrt{d_{\text{head}}}$ & 0.125 & 0.25 & 0.5 & 1.0 (PRISM) & 2.0 & 4.0 & 8.0 \\
\midrule
Retrieve Fruit & 0.22 & 0.54 & 0.54 & 0.64 & 0.64 & 0.54 & 0.62 \\
\bottomrule
\end{tabular}
\caption{Effect of attention sharpness on PRISM performance. We scale attention logits by $\frac{V}{\sqrt{d_{\text{head}}}}$ before softmax; PRISM uses the default setting $1.0$.}
\label{tab:attn_sharpness_ablation}
\end{table*}
We study the effect of shaping the attention distribution affects performance by scaling the attention logits. Making attention too flat ($\text{factor}=0.125$) severely hurts success ($0.22$ vs.\ $0.64$, a $\sim$42\% drop), indicating that overly permissive information flow leads to noisy retrieval.
In contrast, moderate increases in sharpness around the PRISM setting (factors $2.0$--$8.0$) cause mild degradation ($2$--$10$ pp), suggesting that performance is much more sensitive to insufficient selectivity than to conservative selection. Overall, these results indicate that controlling information flow is crucial for performance, a function implemented by PRISM’s gating module and learned directly from data (\autoref{tab:attn_sharpness_ablation}).
\section{Effect of Attention Dropping}
\begin{table*}[ht!]
\centering
\small
\setlength{\tabcolsep}{6pt}
\begin{tabular}{lccccc}
\toprule
Method & Spatial & Prospective & Associative & Object-Set & Average \\
\midrule
PRISM w/o Gating              & 0.08 & 0.35 & \textbf{0.30} & 0.30 & 0.26 \\
PRISM w/o Gating (Attn Drop)  & 0.44 & \textbf{0.80} & 0.10 & 0.20 & 0.39 \\
PRISM                         & 0.40 & \textbf{0.80} & 0.25 & 0.35 & 0.45 \\
PRISM (Attn Drop)             & \textbf{0.60} & 0.64 & \textbf{0.30} & \textbf{0.40} & \textbf{0.49} \\
\bottomrule
\end{tabular}
\caption{Effect of noisy attention (token dropping with probability $p=0.1$) on success rates across the four ReMemBench categories. While noise improves both gated and ungated models, PRISM with learned gating remains strongest overall.}
\label{tab:noisy_attention_gating}
\end{table*}
We also examine whether adding noise to attention can improve robustness. We inject noise by randomly dropping tokens with probability $p=0.1$ [7]. This boosts PRISM’s average success by $+4\%$ ($0.45 \rightarrow 0.49$). Applying the same token-drop strategy to PRISM without gating also helps ($0.26$ $\rightarrow$ $0.39$), but the full PRISM architecture remains stronger overall: it outperforms the no-gating variant by +10\% with attention dropping ($0.49$ vs.\ $0.39$) and +19\% without it ($0.45$ vs.\ $0.26$). These results suggest that noisy attention alone is helpful but insufficient, and that the learned gating module remains crucial for suppressing irrelevant features from retrieval (\autoref{tab:noisy_attention_gating}).

\section{Effect of Gating Module}
\begin{table*}[h]
\centering
\small
\setlength{\tabcolsep}{6pt}
\begin{tabular}{lccccc}
\toprule
Method & Spatial & Prospective & Associative & Object-Set & Average \\
\midrule
Blind Gating          & 0.35 & 0.30 & 0.25 & 0.20 & 0.28 \\
Feature Gating (PRISM) & \textbf{0.40} & \textbf{0.80} & \textbf{0.25} & \textbf{0.35} & \textbf{0.45} \\
\bottomrule
\end{tabular}
\caption{Blind vs.\ feature-wise gating. Blind gating learns a single scalar gate per layer with similar non-linearity and parameter count, yet feature-wise gating in PRISM achieves substantially higher average success.}
\label{tab:blind_vs_feature_gating}
\end{table*}
We acknowledge that modifications introduced by $g(x)$, such as additional parameters or non-linearities, could confound the claim that performance gains stem from filtering irrelevant information. To isolate the effect of feature-wise gating from these factors, we compare blind gating (86M parameters, sigmoid non-linearity) with feature-wise gating (90M parameters, sigmoid non-linearity). Blind gating, instead of producing a per-feature gate, learns a single scalar gate per layer and thus matches PRISM in non-linearity while maintaining a similar parameter count (less than $5\%$ extra). The resulting performance gap is substantial: feature-wise gating yields a $+17\%$ improvement in average success, indicating that the gains primarily arise from feature-wise information selection rather than increased capacity or any non-linearity (\autoref{tab:blind_vs_feature_gating}).
\section{Evaluation on Mikasa-Robo}
\begin{figure*}[t]
    \centering
    \includegraphics[width=\textwidth]{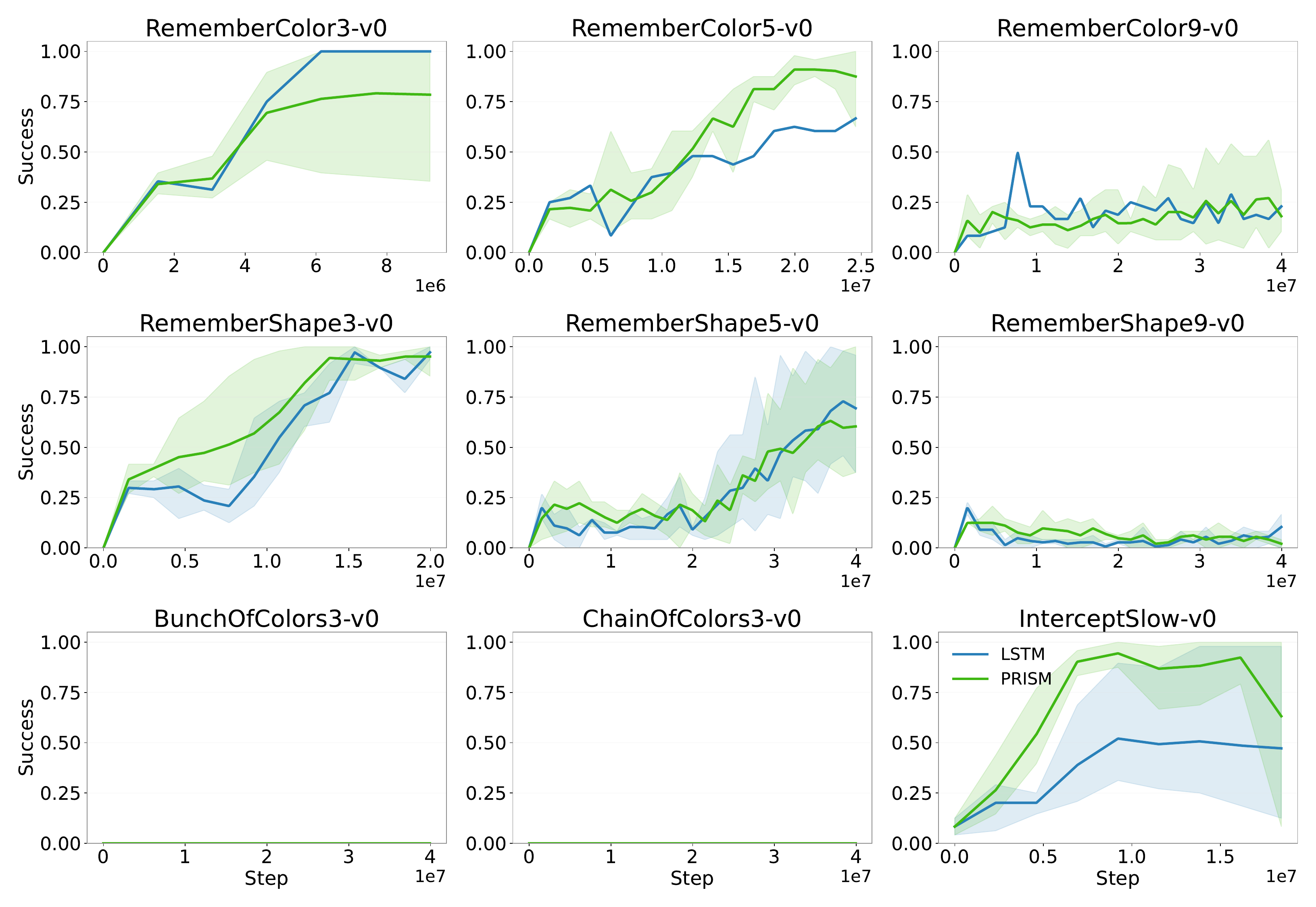}
    \caption{
    Comparison of PRISM vs. LSTM on Mikasa-Robo. 
    }
    \label{fig:mikasa_comparison}
\end{figure*}
\rebuttal{
We also evaluate PRISM on Mikasa-Robo and observe gains on some tasks (e.g., InterceptSlow-v0, RememColor5-v0), but overall performance is comparable to LSTM baselines, unlike on \textsc{ReMemBench}, where transformers clearly outperform LSTMs. We hypothesize this gap stems from differences in task design: Mikasa-Robo tasks have much shorter horizons (180 vs.\ 2.8k timesteps in \textsc{ReMemBench}), contain fewer natural distractors, and concentrate most of the relevant information in the initial frames, making them less diagnostic of robust, long-horizon memory mechanisms.
These differences in performance invite a systematic study of the factors that strongly influence short-term memory use and downstream performance (\autoref{fig:mikasa_comparison}).
}

\section{Analysis of the compression factor ($m$)}\label{sec:app:compression_ratio}
\begin{table*}[t]
\centering
\small
\setlength{\tabcolsep}{6pt}
\begin{tabular}{lccccc}
\toprule
Compression ($m$) & Spatial & Prospective & Associative & Object-Set & Average \\
\midrule
$m \approx 2$             & 0.30 & 0.40 & \textbf{0.28} & 0.24 & 0.31 \\
$m \approx 9$             & \textbf{0.40} & 0.56 & 0.24 & \textbf{0.40} & 0.40 \\
$m \approx 80$            & \textbf{0.40} & 0.60 & 0.24 & \textbf{0.40} & 0.41 \\
$m \approx 200$ (PRISM)   & \textbf{0.40} & \textbf{0.80} & 0.25 & 0.35 & \textbf{0.45} \\
\bottomrule
\end{tabular}
\caption{Effect of compression factor $m$ on success rates across ReMemBench tasks. Larger $m$ corresponds to more aggressive compression.}
\label{tab:compression_ablation}
\end{table*}

\begin{figure}[t]
    \centering
    \includegraphics[width=\linewidth]{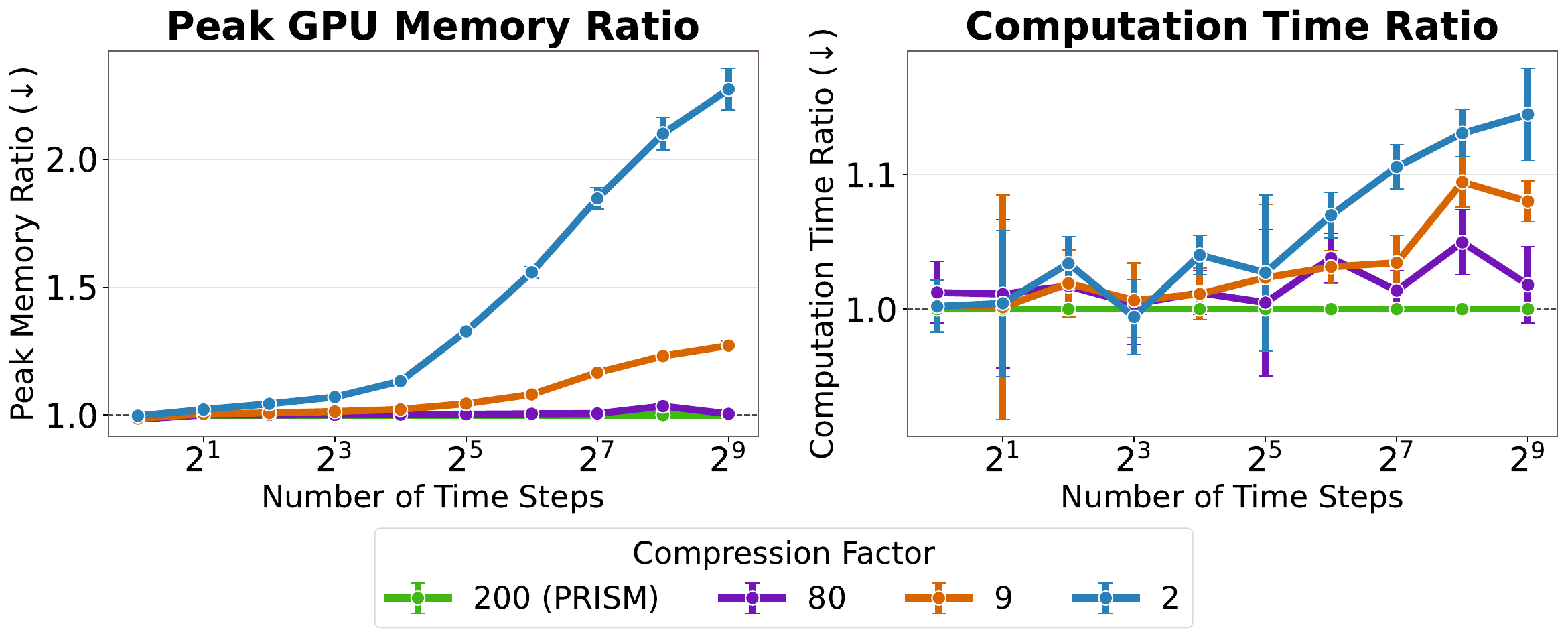}
    \caption{
    Effect of compression factor $m$ on test-time efficiency. 
    The plot reports peak GPU memory and computation time, both normalized with respect to the $m \approx 200$ (PRISM) configuration, across different sequence lengths.
    }
    \label{fig:compression_compute_memory}
\end{figure}
\rebuttal{
The compression factor $m$ trades off memory capacity and resource usage. In our experiments, we found $m \approx 200$ (compressing all tokens from the current timestep) to be a practical operating point under a budget of 384 H100 GPU-hours per experiment; for $m \approx 2$, it was difficult to obtain enough gradient steps to reach reasonable performance. Table~\ref{tab:compression_ablation} reports success rates for $m \approx 2, 9, 80,$ and $200$. As compression becomes more aggressive (larger $m$), we observe a modest but consistent increase in average performance across tasks (from $0.31$ at $m \approx 2$ to $0.45$ at $m \approx 200$), even though individual categories do not follow a strictly monotonic trend. In contrast, the efficiency gains are clear: at sequence length 512, $m \approx 200$ reduces compute by $13\%$ and peak GPU memory by $56\%$ relative to $m \approx 2$ (\autoref{fig:compression_compute_memory}).
}

\section{Memory and Compute Footprint}\label{sec:app:footprint}\label{sec:app:compute_calc}
\paragraph{Training.}
Assuming each local encoder compresses \(k_{\text{tot}}\) tokens to \(k_{\text{tot}}/m\), there are \(n\) local blocks, each processing \(k_{\text{tot}}\) tokens. The local self-attention per block costs \(\mathcal{O}(m^{2})\) and is invoked \(n\,k_{\text{tot}}/m\) times, yielding a total local compute of \(\mathcal{O}(n\,m\,k_{\text{tot}})\). After compression, the global module attends over \(n\,k_{\text{tot}}/m\) summary tokens, incurring \(\mathcal{O}\!\big((n\,k_{\text{tot}}/m)^{2}\big)\). Thus, the overall training-time compute decomposes into a linear-in-\(k_{\text{tot}}\) local term plus a quadratic-in-the-compressed-length global term:
\[
\underbrace{\mathcal{O}(n\,m\,k_{\text{tot}})}_{\text{local}} \;+\; \underbrace{\mathcal{O}\!\big((n\,k_{\text{tot}}/m)^{2}\big)}_{\text{global}}\!.
\]
Training-time \emph{memory} can also benefit from the reduced global length (lesser activation memory), but these savings are less pronounced in practice because modern optimizations (e.g., FlashAttention tiling/splitting) already reduce activation memory, the major contributor to training-time footprint for long-sequence transformers.

\smallskip
\paragraph{Evaluation.}
The same decomposition applies to evaluation compute, but the memory benefit is much clearer. Operating global attention over \(n\,k_{\text{tot}}/m\) summary tokens instead of all \(n\,k_{\text{tot}}\) tokens reduces both global-attention compute and the key-value (KV) cache by a factor of \(\approx m\). Concretely, at inference, the model only attends to \(n\,k_{\text{tot}}/m\) tokens and stores proportionally fewer KV entries, yielding an \(\approx m\times\) reduction in global-attention runtime and memory relative to full/global attention over the uncompressed sequence. Empirical evaluation-time measurements (compute and memory) are reported in the main paper.

\smallskip
\paragraph{Baselines}
To quantify empirical compute and memory savings, we compare PRISM to two baselines: (i) standard full attention over the entire history, and (ii) a hybrid with full global attention plus a sliding window. For each method, we run a unit block \(20\) times, recording wall-clock time and peak memory. We then normalize these measurements by the number of learnable parameters and report PRISM-to-baseline ratios for both compute time and memory.

\section{Implementation Details}\label{sec:app:implementation}
\paragraph{\textsc{PRISM} Architecture Details}
We set $k_p{=}1$, $k_v{=}198 \times 2$ (tokens per camera $\times$ cameras), and $k_a{=}1$. In all experiments, the \emph{local} encoder compresses the $(k_p{+}k_v)$ tokens to a single summary token; thus the per-step token count changes from $(k_p{+}k_v{+}k_a){=}398$ to $(1{+}k_a){=}2$, yielding a compression factor
$m \!=\! \frac{k_p + k_v + k_a}{1 + k_a} \!=\! \frac{398}{2} \!=\! 199$.
We instantiate the Transformer backbone with a standard LLaMA-style implementation. Visual features are extracted using \emph{CrossMAE}~\citep{fu2024rethinking} and kept \emph{frozen} during policy training. The action head is a two-layer MLP.

\paragraph{Data Collection} \rebuttal{For each of the \textsc{ReMemBench} tasks, $50$ expert demonstrations are collected by a human teleoperator using a spacemouse for imitation learning. Similarly, for the real-world tasks, we collect $100$ expert demonstrations using the spacemouse.}

\begin{table}[h]
\centering
\caption{Training hyperparameters.}
\label{tab:training-config}
\resizebox{0.8\columnwidth}{!}{%
\begin{tabular}{ll}
\toprule
\textbf{Config} & \textbf{Value} \\
\midrule
Optimizer & AdamW \\
Base learning rate & $5\times10^{-4}$ \\
Effective batch size & 256 \\
Weight decay & $0.01$ \\
Warmup epochs & $2$ \\
Total epochs & $200$ \\
Action prediction horizon & $32$ \\
Proprioception noise (std) & $0.005$ \\
Brightness augmentation & $\mathrm{Uniform}(-0.1,\,0.1)$ \\
Contrast augmentation & $\mathrm{Uniform}(0.8,\,1.2)$ \\
Number of cameras & $2$ \\
\bottomrule
\end{tabular}
}
\end{table}
\paragraph{Training Details}\label{sec:app:training}
We train for all eight \textsc{ReMemBench} tasks using \(\sim\)384 H100 GPU-hours, and for four \textsc{RoboCasa} atomic tasks using \(\sim\)192 H100 GPU-hours.

\end{document}